%% file: samplepaper.tex
\documentclass[12pt]{article}

\usepackage{sbc-template}

\usepackage{graphicx,url}
\usepackage[utf8]{inputenc}  
\usepackage[T1]{fontenc}
\usepackage{tcolorbox}
\tcbuselibrary{breakable}
\usepackage{float}
\usepackage{booktabs}
\title{\textsc{Check-Eval}: A Checklist-based Approach for Evaluating Text Quality}
\author{Jayr Pereira\inst{1,2} \and Andre Assumpcao \inst{3} \and
Roberto Lotufo\inst{2,4}}
\address{Universidade Federal do Cariri (UFCA), Juazeiro do Norte-CE, Brazil
\nextinstitute
Universidade Estadual de Campinas (UNICAMP), Campinas-SP, Brazil
\nextinstitute
National Center for State Courts (NCSC), Williamsburg, Virginia, United States
\nextinstitute
NeuralMind.ai, Brazil
  \email{\{jap2, snm, cz, rdnf\}@cin.ufpe.br}
}

\begin{document}
\maketitle              
\begin{abstract}
Evaluating the quality of text generated by large language models (LLMs) remains a significant challenge. Traditional metrics often fail to align well with human judgments, particularly in tasks requiring creativity and nuance. In this paper, we propose \textsc{Check-Eval}, a novel evaluation framework leveraging LLMs to assess the quality of generated text through a checklist-based approach. \textsc{Check-Eval} can be employed as both a reference-free and reference-dependent evaluation method, providing a structured and interpretable assessment of text quality. The framework consists of two main stages: checklist generation and checklist evaluation. We validate \textsc{Check-Eval} on two benchmark datasets: Portuguese Legal Semantic Textual Similarity and \textsc{SummEval}. Our results demonstrate that \textsc{Check-Eval} achieves higher correlations with human judgments compared to existing metrics, such as \textsc{G-Eval} and \textsc{GPTScore}, underscoring its potential as a more reliable and effective evaluation framework for natural language generation tasks. The code for our experiments is available at \url{https://anonymous.4open.science/r/check-eval-0DB4}.


\end{abstract}

\section{Introduction} \label{sec:introduction}

Evaluating the quality of text generated by large language models (LLMs) remains an open problem within the field of natural language generation (NLG). Traditional evaluation metrics, such as BLEU \cite{papineni2002bleu}, ROUGE \cite{lin2004rouge}, and METEOR \cite{banerjee2005meteor}, often show limited correlation with human judgments, especially in tasks that require creativity and diversity, such as dialogue generation and summarization. Despite advancements in LLMs, which can produce high-quality, fluent texts that closely mirror human writing, the challenge lies in accurately assessing these outputs quality.

Recent approaches utilizing LLMs as text quality evaluators have shown promise, yet they still fall short in achieving reliable alignment with human judgments \cite{liu2023geval,fu2023gptscore}. This ongoing challenge underscores the necessity for more effective and scalable evaluation frameworks that can bridge the gap between automated metrics and human evaluators, ensuring that the outputs of NLG systems meet the desired standards of coherence, relevance, and overall quality \cite{liu2023geval}.

In this paper, we propose \textsc{Check-Eval}, a text evaluation framework that leverages the strengths of LLMs to assess the quality of generated text. The proposed method instructs the LLM to generate a checklist of key points that should be present in a candidate text for it to be considered high-quality. This checklist is derived from the source or reference text, providing a structured and interpretable reference for evaluating the candidate. By comparing the candidate text to the generated checklist, \textsc{Check-Eval} can provide a nuanced and comprehensive assessment of text quality, capturing essential elements such as content consistency, coherence, and relevance.


We evaluate \textsc{Check-Eval} in two main scenarios, both based on human judgments: (1) the Portuguese Legal Semantic Textual Similarity dataset \cite{silvajr2024dataset}, a benchmark dataset for evaluating the semantic similarity of legal texts in Portuguese, and (2) the \textsc{SummEval} dataset \cite{fabbri2021summeval}, a benchmark dataset for text summarization evaluation. Our experiments demonstrate that \textsc{Check-Eval} achieves higher correlations with human judgments compared to existing metrics, such as \textsc{G-Eval} and \textsc{GPTScore}, highlighting its potential as a more reliable and effective evaluation framework for NLG tasks. Additionally, we show that \textsc{Check-Eval} can identify specific areas of improvement in the generated summaries, providing valuable feedback for model development and refinement.

The main contributions of this paper are:
\begin{itemize}
    \item \textbf{Introduction of \textsc{Check-Eval}}: A novel evaluation framework leveraging LLMs for text quality assessment.
    \item \textbf{Comprehensive Evaluation}: Demonstrating the superior performance of \textsc{Check-Eval} on two benchmark datasets.
    \item \textbf{Insightful Feedback}: Highlighting the ability of \textsc{Check-Eval} to identify specific areas of improvement, aiding in model development and refinement.
\end{itemize}

The remainder of this paper is organized as follows: Section \ref{sec:related_work} provides an overview of related work. Section \ref{sec:check_eval} introduces the \textsc{Check-Eval} framework, detailing the checklist generation and evaluation stages. Section \ref{sec:experiments} describes the experimental settings and dataset used to evaluate \textsc{Check-Eval}. Section \ref{sec:results} presents the results of the experiments, comparing \textsc{Check-Eval} to existing metrics. Finally, Section \ref{sec:conclusions} concludes the paper and discusses future directions for research.

\section{Related Work} \label{sec:related_work}

The evaluation of automatically generated text has been a persistent challenge in the NLG field. Traditional metrics such as BLEU, ROUGE, and METEOR have been extensively used but have shown limitations in aligning with human judgment, particularly in tasks requiring creativity and nuance \cite{fabbri2021summeval}. In recent years, more sophisticated evaluation frameworks leveraging LLMs have been proposed to address these shortcomings. Two recent methods are \textsc{GPTScore} \cite{fu2023gptscore} and \textsc{G-Eval} \cite{liu2023geval}.

\textsc{GPTScore} is a framework that utilizes generative pre-trained transformers (GPTs) and other language models to evaluate NLG outputs without relying on reference texts. The core idea is to assess the probability that the LLM assigns to the generated text, under the assumption that higher probabilities indicate higher quality. Fu et al. (2023) \cite{fu2023gptscore} demonstrated that \textsc{GPTScore} could achieve better correlations with human judgments compared to traditional metrics, especially in open-ended tasks such as dialogue generation and creative writing. However, despite its advancements, the method lacks interpretability and may be biased towards texts similar to those seen during the model's training phase.

\textsc{G-Eval} is another recent approach that leverages the capabilities of LLMs, specifically GPT-4, to improve NLG evaluation. Proposed by Liu et al. (2023) \cite{liu2023geval}, \textsc{G-Eval} introduces a chain-of-thought (CoT) \cite{wei2022cot} paradigm where the evaluation process is guided by detailed intermediate steps generated by the LLM. This method has shown improvements in correlation with human evaluations, particularly in tasks such as text summarization and dialogue generation. To address issues with score distribution and variability, \textsc{G-Eval} employs a self-consistency strategy. Specifically, it generates multiple samples ($n = 20$) using different decoding parameters and averages the evaluation scores across these samples. This approach helps mitigate two key problems: the dominance of a single score (such as 3 on a 1-5 scale) and the tendency of LLMs to output integer scores, even when decimal values are requested. By using self-consistency, \textsc{G-Eval} captures more subtle differences between generated texts and provides a more reliable assessment of text quality.

While both \textsc{GPTScore} and \textsc{G-Eval} represent significant advancements in automatic text quality evaluation, \textsc{Check-Eval} aims to address their limitations through a novel checklist-based approach. Unlike \textsc{GPTScore}, which relies on the probabilistic output of LLMs, \textsc{Check-Eval} generates a checklist of key points derived from the source text. This checklist serves as a concrete reference against which the generated summary is evaluated, providing a more structured and interpretable assessment of content consistency, coherence, and relevance.

Furthermore, \textsc{Check-Eval} builds upon the structured evaluation paradigm of \textsc{G-Eval} but simplifies the evaluation process by focusing on key content points rather than generating comprehensive CoT steps. This approach not only reduces the potential for bias towards LLM-generated texts but also enhances the scalability and applicability of the evaluation framework to a wider range of NLG tasks. By providing actionable feedback, \textsc{Check-Eval} can more effectively guide model improvement and refinement, ultimately leading to the generation of higher-quality text.

\section{\textsc{Check-Eval}} \label{sec:check_eval}

In this section, we introduce \textsc{Check-Eval}, an LLM-based evaluation framework designed to assess the quality of automatically generated text. \textsc{Check-Eval} leverages the capabilities of large language models (LLMs) to generate and evaluate a checklist of key points derived from the reference document and the candidate text. The primary aim of \textsc{Check-Eval} is to provide a more structured, interpretable, and comprehensive assessment of text quality.

\begin{figure}[t]
\centering
\includegraphics[width=0.8\textwidth]{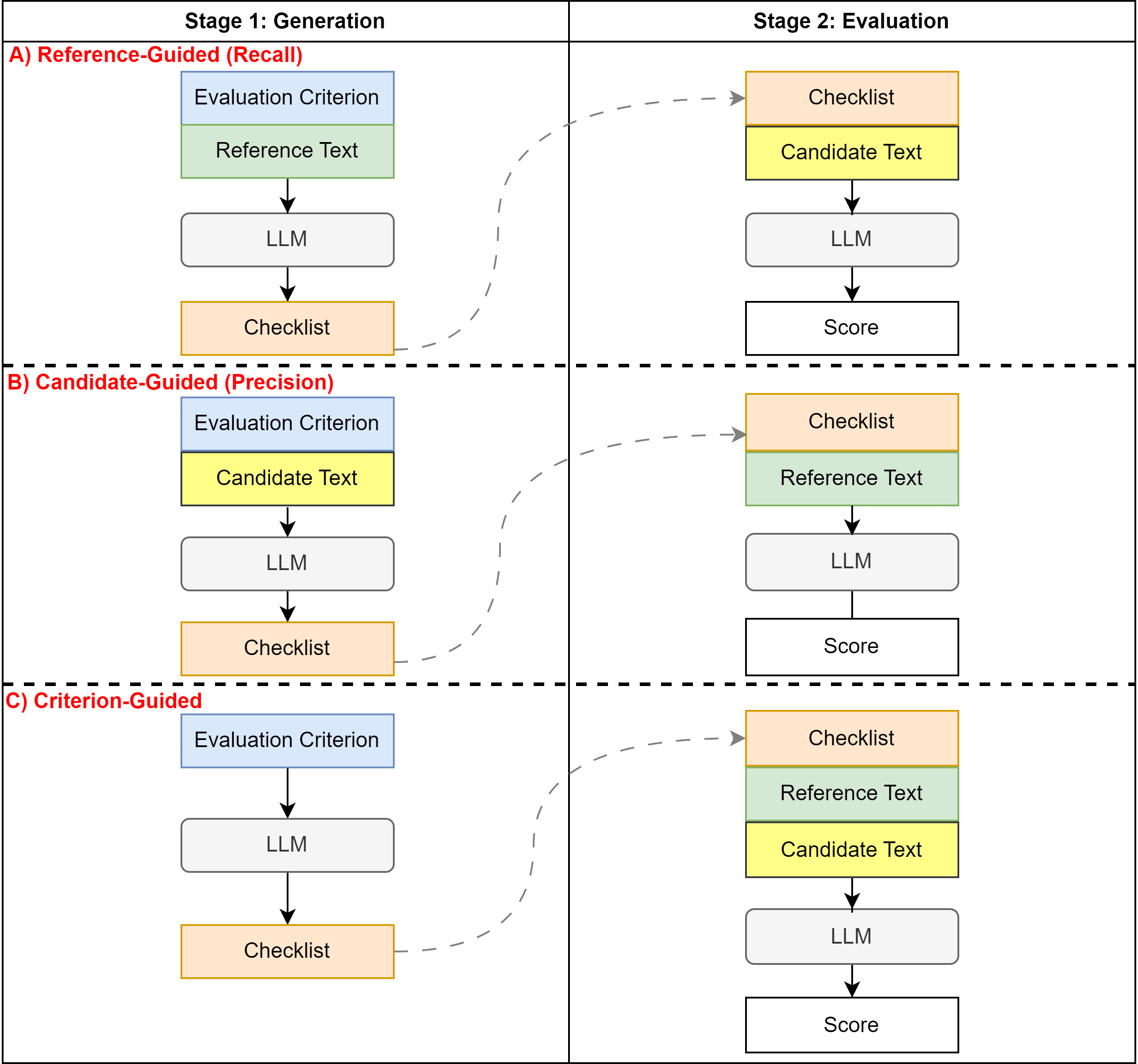}
\caption{Illustration of the \textsc{Check-Eval} methodology.}
\label{fig:check_eval}
\end{figure}

The \textsc{Check-Eval} framework consists of two main stages: checklist generation and checklist evaluation, as illustrated in Figure \ref{fig:check_eval}. The framework has three main variations: (1) Reference-Guided, (2) Candidate-Guided, and (3) Criterion-Guided evaluation. The Reference-Guided variation uses a reference text to generate the checklist, which is then used to evaluate the candidate text. The Candidate-Guided variation works similarly, but the checklist is generated from the candidate text and used to evaluate the reference text. These two variations function as recall and precision evaluations, respectively.

The Criterion-Guided variation uses specific evaluation criteria to generate the checklist, which is then used to evaluate the candidate text against the reference text. It is useful when no reference text is available, such as in the case of text summarization tasks. Deciding which variation to use depends on the evaluation scenario and the characteristics of the dataset. The following sections describe the checklist generation and evaluation stages in detail.

\subsection{Checklist Generation}

The first stage of \textsc{Check-Eval} involves generating a checklist of key points to evaluate the quality of the candidate text. Below, we describe the checklist generation process for the framework's variations.

\subsubsection{Reference-Guided and Candidate-Guided Checklist Generation}

For this step, let's consider the reference and the candidate text used in the reference-guided and candidate-guided variations, respectively, as the source document. The LLM is prompted to extract the essential information from the source document and create an evaluation checklist based on predefined evaluation criteria. The checklist serves as a structured reference for the key points that should be present in the text to be evaluated.

\begin{figure}[t]
\begin{tcolorbox}
\textbf{[System]}\\
Your task is to write a checklist of the elements a \textcolor{blue}{relevant} text summary should have.\\
\\
\textcolor{blue}{Relevance definition: selection of important content from the source. The summary should include only important information from the source document. Annotators were instructed to penalize summaries that contained redundancies and excess information.}\\
\\
You must follow the following rules:\\
\\
1. The checklist should contain yes/no questions.\\
2. Checklist items must be self-contained.\\
3. Focus on the main concepts and avoid including minor details.\\
4. Each item should represent a unique concept or element from the text.\\
5. Avoid repetition and overlapping of concepts in the checklist items.\\
6. The checklist should be comprehensive but not exhaustive, aiming for clarity and brevity.\\
7. Generate as many items as you think are necessary\\

\textbf{[User]}
\\
\\
\{source text\}
\end{tcolorbox}
\caption{Prompt used to generate the checklist from a source text. The blue text is the definition of the evaluation criteria, which is a variable that can be changed according to the desired evaluation criteria (e.g., consistency, coherence, relevance and fluency).}
\label{fig:prompt_generate_checklist}
\end{figure}

Figure \ref{fig:prompt_generate_checklist} shows the prompt used to generate the checklist from a source text given specific evaluation criteria. The criteria can be adjusted according to the desired evaluation focus, such as consistency, coherence, relevance, and fluency, which are generally used in text summarization tasks. Defining the evaluation criteria is crucial to ensure that the generated checklist captures the essential elements that should be present in a high-quality summary. The checklist generation process is repeated for each source document and each evaluation criterion, resulting in a set of reference checklists that can be used to evaluate candidate texts. The prompt is designed to guide the LLM in generating a comprehensive and relevant checklist that captures the main concepts and elements of the source document, avoiding redundancies and minor details.

The output of the checklist generation stage is a list of yes/no questions that represent the key points extracted from the source document. Each question corresponds to a unique concept or element from the text that should be present in a high-quality candidate text. 

To illustrate the interpretability power of \textsc{Check-Eval} evaluation, consider the example of a generated checklist based on a source document about climate change shown in Figure \ref{fig:checklist_example}. This example demonstrates how \textsc{Check-Eval} generates a structured and interpretable checklist that captures the key points of the source document that should be present in a high-quality candidate text.

\begin{figure}[t]
\begin{tcolorbox}
\textbf{Source Document Excerpt:} "Climate change refers to long-term shifts and alterations in temperature and weather patterns. These changes may be natural, such as through variations in the solar cycle. However, since the 1800s, human activities have been the main driver of climate change, primarily due to the burning of fossil fuels like coal, oil, and gas."\\

\textbf{Generated Checklist:}
\begin{enumerate}
\item Does the summary mention that climate change refers to long-term shifts in temperature and weather patterns?
\item Does the summary state that these changes can be natural, such as variations in the solar cycle?
\item Does the summary highlight that human activities have been the main driver of climate change since the 1800s?
\item Does the summary specify that burning fossil fuels like coal, oil, and gas is a primary cause?
\\...
\end{enumerate}
\end{tcolorbox}
\caption{Example of a generated checklist based on a source document about climate change. The checklist aims to capture the key points of the source document and serves as a reference for evaluating the candidate summary.}
\label{fig:checklist_example}
\end{figure}

\subsubsection{Criterion-Guided Checklist Generation}

In the criterion-guided variation, the checklist is generated based on specific evaluation criteria, such as consistency, coherence, relevance, and fluency. The LLM is prompted to create a checklist of key points to consider when evaluating the quality of a text in comparison to the reference text, based on the chosen evaluation criteria. The prompt used for this variation is similar to the one used for the reference-guided and candidate-guided variations, but it only includes the evaluation criteria, as shown in Figure \ref{fig:prompt_evaluate_summary}. The checklist generated in this variation is used to evaluate the candidate text against the reference text.

\subsection{Checklist Evaluation}

The second stage of the \textsc{Check-Eval} framework involves evaluating the candidate text based on the generated checklist. The LLM is prompted to compare the content of the candidate text to the key points in the checklist and determine the presence or absence of each key point. 

\begin{figure}[t]
\begin{tcolorbox}
\textbf{[System]}\\
Task: assess the consistency of a summary of a source text.
\\
Consistency definition: the factual alignment between the summary and the summarized source. A factually consistent summary contains only statements that are entailed by the source document. Annotators were also asked to penalize summaries that contained hallucinated facts.
\\
\\
Checklist:
\\
\\
\{generated checklist\}
\\
\\
\textbf{[User]}
\\
\{candidate summary\}
\end{tcolorbox}
\caption{Prompt used to evaluate a candidate summary based on the generated checklist. The checklist is specific to the evaluation criteria, in this case, consistency.}
\label{fig:prompt_evaluate_summary}
\end{figure}

Figure \ref{fig:prompt_evaluate_summary} shows the prompt used to evaluate a candidate text based on the generated checklist for either the reference-guided or candidate-guided variations. The prompt specifies the evaluation criteria—in this case, consistency—which is an optional variable that can be changed according to the desired evaluation focus. The LLM is instructed to assess the defined criteria of the text by comparing its content to the key points in the checklist and determining whether each key point is present. The checklist ensures a structured and comprehensive assessment.

For the criterion-guided variation, the prompt is similar to the one used for the reference-guided and candidate-guided variations, but the reference and the candidate text are passed together with the checklist generated based on the chosen evaluation criteria. The LLM is prompted to evaluate the candidate text against the reference text based on the checklist.

The output of the checklist evaluation stage is a score that reflects the quality of the candidate text based on the presence or absence of key points in the checklist. The score indicates how well the text captures the essential elements of the source document according to specific evaluation criteria. For example, in the case of evaluating consistency, the score reflects the factual alignment between the reference and the candidate text, considering the presence of entailed statements and the absence of hallucinated facts. The final score is calculated by counting the number of key points present in the text, providing a quantitative measure of the text's quality.

\section{Experiments} \label{sec:experiments}

In this section, we describe the experiments conducted to evaluate the performance of \textsc{Check-Eval}. We begin by detailing the datasets used, followed by the experimental settings.

\subsection{Portuguese Legal Semantic Textual Similarity}

We evaluated \textsc{Check-Eval} using the Portuguese Legal Semantic Textual Similarity \cite{silvajr2024dataset} dataset, a benchmark dataset for evaluating the semantic similarity of legal texts in Portuguese. The dataset consists of pairs of legal documents, each annotated with a similarity score. The dataset provides two versions of the annotations: one performed by legal experts on 32 pairs of legal documents and another automated annotation using heuristics based on legal cases metadata. The overall correlation between the human and automated annotations is 0.45 (Pearson correlation coefficient) and 0.43 (Spearman correlation coefficient). We experimented with the 32 pairs annotated by legal experts and a subset of 100 pairs annotated by the automated method randomly selected from the dataset.

\subsubsection{Experimental Settings}

We evaluated the first two variations of \textsc{Check-Eval} (Reference-Guided and Candidate-Guided) using the Portuguese Legal Semantic Textual Similarity dataset. As we consider the reference-guided variation as the recall evaluation and the candidate-guided variation as the precision evaluation, we also computed the F1 score as the harmonic mean of the recall and precision scores. We used OpenAI's GPT-4-turbo model to perform both checklist generation and checklist evaluation. The following steps outline the experimental setup:

\begin{enumerate}
    \item \textbf{Checklist Generation:} For each text pair (text 1 and text 2) in the dataset, we prompted the LLM to generate a checklist of key points based on the criteria of relevance, coherence, consistency, and fluency. For the reference-guided variation, the checklist was generated from text 1, and for the candidate-guided variation, the checklist was generated from text 2. The prompt used for checklist generation is a translation of the prompt used for the \textsc{SummEval} dataset, as shown in Figure \ref{fig:prompt_generate_checklist}.
    \item \textbf{Checklist Evaluation:} Each text pair was evaluated based on the generated checklist. The LLM was prompted to assess the presence or absence of each key point in text 2 (candidate text) compared to text 1 (reference text). The prompt used for checklist evaluation is a translation of the prompt used for the \textsc{SummEval} dataset, as shown in Figure \ref{fig:prompt_evaluate_summary}.
    \item \textbf{Scoring:} The final score for each text pair was calculated by counting the number of key points present in text 2, providing a quantitative measure of its quality. The F1 score was computed as the harmonic mean of the recall and precision scores.
\end{enumerate}

\subsection{\textsc{SummEval} Dataset}

We also evaluated \textsc{Check-Eval} in a reference-free evaluation scenario using the \textsc{SummEval} dataset \cite{fabbri2021summeval}. The \textsc{SummEval} dataset consists of automatically generated summaries for news articles from the CNN/DailyMail dataset, along with human annotations for the quality dimensions of coherence, consistency, fluency, and relevance. This dataset provides a comprehensive benchmark for comparing the performance of different evaluation metrics against human judgments. In this case, we evaluated the Criterion-Guided variation of \textsc{Check-Eval} using the \textsc{SummEval} dataset.

\subsubsection{Experimental Settings}

For our experiments, we used the GPT-4 model to perform both checklist generation and checklist evaluation. We prompted GPT-4 to generate a checklist of key points based on the evaluation criteria provided in the \textsc{SummEval} paper \cite{fabbri2021summeval}. The following steps outline the experimental setup:

\begin{enumerate}
    \item \textbf{Checklist Generation:} For each of the evaluation criteria (consistency, relevance, coherence, and fluency), we prompted GPT-4 to generate a checklist of key points based on the criteria definitions provided in the \textsc{SummEval} paper \cite{fabbri2021summeval}.
    
    \item \textbf{Checklist Evaluation:} Each candidate summary was evaluated against the generated checklist. GPT-4 was prompted to assess the presence or absence of each key point in the summary. The prompt used for checklist evaluation is shown in Figure \ref{fig:prompt_evaluate_summary}.

    \item \textbf{Scoring:} The final score for each candidate summary was calculated by counting the number of key points present in the summary, providing a quantitative measure of its quality.
\end{enumerate}

\section{Results} \label{sec:results}

In this section, we present the results of our experiments evaluating the performance of \textsc{Check-Eval}.

\subsection{Portuguese Legal Semantic Textual Similarity}

Table \ref{tab:check_eval_pt} summarizes the results of \textsc{Check-Eval} on the Portuguese Legal Semantic Textual Similarity dataset. The table presents the Pearson ($\rho$) and Spearman ($\rho_s$) correlations for the reference-guided and candidate-guided variations, as well as the F1 score, which is the harmonic mean of the recall (reference-guided) and precision (candidate-guided) scores. The results demonstrate that \textsc{Check-Eval} achieves correlation scores above the 0.45 (Pearson) and 0.43 (Spearman) of the automated annotations from the dataset, indicating that \textsc{Check-Eval} provides a reliable and effective evaluation of text similarity in the legal domain.

\input{tables/check_eval_pt}

We evaluated the performance of \textsc{Check-Eval} across four evaluation criteria: consistency, relevance, coherence, and fluency. Additionally, we computed the average correlation across all criteria to provide an overall assessment of \textsc{Check-Eval}'s performance. The results show that \textsc{Check-Eval} achieves competitive correlations with human judgments across all criteria. The F1 score also demonstrates the effectiveness of \textsc{Check-Eval} in capturing the essential elements of the source text in the candidate summary.

The candidate-guided variation of \textsc{Check-Eval} achieves higher correlations with human judgments compared to the reference-guided variation. However, for this dataset, the F1 score provides a more balanced evaluation of text quality, as both candidate and reference texts are equally important. Overall, the results indicate that \textsc{Check-Eval} is a reliable and effective evaluation framework for assessing the quality of text similarity in the legal domain.

We also evaluated \textsc{Check-Eval} using the automated annotations from the dataset. Table \ref{tab:check_eval_pt_heristics} presents the results for a sub-sample of 100 pairs automatically annotated by the heuristics proposed in \cite{silvajr2024dataset}. The results show that \textsc{Check-Eval} achieves higher correlations with the automated annotations compared to the human annotations. For example, the reference-guided variation shows a Pearson correlation of 0.53 for consistency with automated annotations, whereas it shows 0.45 with human annotations. This suggests that the automated annotations align more closely with the systematic evaluation approach of \textsc{Check-Eval}.

\input{tables/check_eval_pt_heristic}

The higher correlations with automated annotations could be attributed to the nature of automated heuristics, which tend to be more consistent and systematic compared to human annotations that can be subjective and vary significantly between annotators. Automated annotations might follow strict rules that closely match the checklist-based approach of \textsc{Check-Eval}, resulting in higher alignment.

Moreover, the fluency criterion was excluded from the evaluation with automated annotations as it showed negative correlations at early evaluation stages. This exclusion might have contributed to the higher average correlations observed with automated annotations, as fluency tends to be more subjective and harder to evaluate consistently by both automated methods and LLMs.

\subsection{\textsc{SummEval} Dataset}

We compare \textsc{Check-Eval} against existing reference-free evaluation metrics, including \textsc{G-Eval} and \textsc{GPTScore}, using the \textsc{SummEval} dataset. The primary evaluation metrics used are Spearman's rank correlation coefficient ($\rho$) and Kendall-Tau correlation coefficient ($\tau$), which measure the correlation between the automated evaluation scores and human judgments.
Table \ref{tab:check_eval} summarizes the results of \textsc{Check-Eval} compared to \textsc{G-Eval} and \textsc{GPTScore} across four evaluation criteria: consistency, relevance, coherence, and fluency. Additionally, we report the average correlation across all criteria.

\input{tables/check_eval}

\subsubsection{Consistency}

\textsc{Check-Eval} achieves a Spearman correlation of 0.605 and a Kendall-Tau correlation of 0.570 for the consistency criterion, outperforming both \textsc{G-Eval} (0.507, 0.425) and \textsc{GPTScore} (0.449, -) significantly. This indicates that \textsc{Check-Eval}'s checklist-based approach provides a more reliable assessment of the factual alignment between the generated summaries and the source documents.

\subsubsection{Relevance}

For relevance, \textsc{Check-Eval} achieves a Spearman correlation of 0.502 and a Kendall-Tau correlation of 0.420. Although \textsc{G-Eval} slightly outperforms \textsc{Check-Eval} in this criterion with a Spearman correlation of 0.547 and Kendall-Tau correlation of 0.433, \textsc{Check-Eval} still shows competitive performance, highlighting its robustness across different evaluation dimensions.

\subsubsection{Coherence}

\textsc{Check-Eval} demonstrates strong performance in coherence, with a Spearman correlation of 0.563 and a Kendall-Tau correlation of 0.461, compared to \textsc{G-Eval}'s 0.582 and 0.457. Although \textsc{G-Eval} slightly surpasses \textsc{Check-Eval} in this criterion, the results indicate that \textsc{Check-Eval}'s structured approach to evaluating the logical flow and clarity of summaries is highly effective.

\subsubsection{Fluency}

For fluency, \textsc{Check-Eval} achieves the highest Spearman and Kendall-Tau correlations (0.490 and 0.446, respectively), outperforming \textsc{G-Eval} (0.455, 0.378) and \textsc{GPTScore} (0.403). This suggests that \textsc{Check-Eval} is particularly adept at assessing the grammatical and stylistic quality of the generated text.

\subsubsection{Overall Performance}

Overall, \textsc{Check-Eval} achieves the highest average correlations across all criteria. These results were computed by averaging the human annotation scores for each criterion and comparing them to the automated evaluation scores generated by \textsc{Check-Eval}, \textsc{G-Eval}, and \textsc{GPTScore}. \textsc{Check-Eval} demonstrates superior performance with an average Spearman correlation of 0.623 and an average Kendall-Tau correlation of 0.493. These results demonstrate that \textsc{Check-Eval} provides a more comprehensive and accurate evaluation of generated text quality compared to existing metrics.

The superior performance of \textsc{Check-Eval} can be attributed to its checklist-based approach, which allows for a more detailed and structured evaluation of key content points. Traditional metrics often struggle with subjective aspects of text quality, such as coherence and relevance, due to their reliance on surface-level comparisons. In contrast, \textsc{Check-Eval} systematically identifies and evaluates essential elements within the text, ensuring that all critical aspects are considered.

\section{Conclusions} \label{sec:conclusions}

In this paper, we introduced \textsc{Check-Eval}, a novel evaluation framework for assessing the quality of automatically generated text. \textsc{Check-Eval} leverages large language models (LLMs) to generate and evaluate a checklist of key points derived from the source document and the candidate summary. Our experiments demonstrated that \textsc{Check-Eval} significantly outperforms existing text evaluation metrics, such as \textsc{GPTScore} and \textsc{G-Eval}, in terms of correlation with human judgments across various dimensions of text quality, including consistency, relevance, coherence, and fluency.

One of the key strengths of \textsc{Check-Eval} is its ability to provide a structured and interpretable assessment of text quality. By focusing on key points extracted from the source document, \textsc{Check-Eval} offers a detailed evaluation framework that aligns with human judgment. The checklist-based approach aims to ensure that all essential elements of the source document are considered, leading to a more comprehensive and focused evaluation. This way, this method tries to mitigate the ambiguity and variability often associated with human judgment by standardizing the evaluation criteria and providing a concrete framework for assessment.

Additionally, \textsc{Check-Eval} reduces the potential for bias present in probabilistic models like \textsc{GPTScore}, which may favor texts similar to those seen during the model's training phase. By concentrating on the presence of specific content elements, \textsc{Check-Eval} provides a fairer and more objective evaluation. This approach also offers actionable feedback by pinpointing specific areas where the generated text deviates from expected quality standards, thus aiding in targeted improvements and refinements.

We evaluated \textsc{Check-Eval} using the Portuguese Legal Semantic Textual Similarity dataset, where it demonstrated higher correlation scores with human judgments compared to automated annotations, indicating its reliability and effectiveness in the legal domain. We also evaluated our method using the \textsc{SummEval} dataset. In this case, the results indicate that \textsc{Check-Eval} achieves higher correlations with human judgments compared to traditional metrics and other LLM-based evaluators. This superior performance underscores the potential of \textsc{Check-Eval} as a reliable and effective evaluation framework for NLG tasks.

Despite its strengths, our study has some limitations. Firstly, the performance of \textsc{Check-Eval} is inherently tied to the capabilities of the underlying LLM, which may introduce biases or errors in checklist generation and evaluation. Secondly, the current implementation of \textsc{Check-Eval} may require significant computational resources, making it less accessible for researchers with limited resources. Lastly, while we demonstrated the effectiveness of \textsc{Check-Eval} in the context of text summarization and legal text similarity, its performance in other NLG tasks remains to be thoroughly evaluated.

Future work should focus on addressing these limitations by exploring ways to optimize the computational efficiency of \textsc{Check-Eval} and extending its application to a broader range of NLG tasks. Additionally, further research is needed to refine the checklist generation process to minimize potential biases and improve the robustness of evaluations.

\section*{Declaration of Generative AI and AI-assisted technologies in the writing process}

During the preparation of this paper, the authors used ChatGPT to check the grammar and semantics of the human written text. After using this tool/service, the authors reviewed and edited the content as needed and take full responsibility for the content of the publication.

%
%
%
\bibliographystyle{sbc}
\bibliography{mybibliography}
\end{document}

%% file: tables/check_eval_pt.tex
\begin{table}[t]
\centering
\caption{\textsc{Check-Eval} results for Portuguese Legal Semantic Textual Similarity. The table presents ($\rho$) Pearson and ($\rho_s$) Spearman correlations.}
\label{tab:check_eval_pt}
\begin{tabular}{@{}lcccccccccc@{}}
\toprule
Method     & \multicolumn{2}{c}{Consistency} & \multicolumn{2}{c}{Relevance} & \multicolumn{2}{c}{Coherence} & \multicolumn{2}{c}{Fluency} & \multicolumn{2}{c}{AVG} \\ 
           & $\rho$        & $\rho_s$       & $\rho$       & $\rho_s$      & $\rho$       & $\rho_s$      & $\rho$      & $\rho_s$     & $\rho$    & $\rho_s$   \\ \midrule
Reference-guided & 0.45          & 0.52          & 0.58         & 0.63         & 0.49         & 0.57         & 0.16        & 0.139        & 0.56        & 0.61        \\
Candidate-guided & 0.55          & 0.62          & 0.58         & 0.59         & 0.55         & 0.55         & 0.25        & 0.14        & 0.62        & 0.62        \\
F1                  & 0.47          & 0.51          & 0.58         & 0.58         & 0.51         & 0.53         & 0.28        & 0.17       & 0.58        & 0.59        \\ \bottomrule
\end{tabular}

\end{table}

%% file: tables/check_eval_pt_heristic.tex
\begin{table}[b]
\centering
\caption{\textsc{Check-Eval} results for Portuguese Legal Semantic Textual Similarity. The table presents ($\rho$) Pearson and ($\rho_s$) Spearman correlations computed for a sub-sample of 100 pairs automatically annotated by the heuristics proposed in \cite{silvajr2024dataset}.}    
\label{tab:check_eval_pt_heristics}
\begin{tabular}{@{}lcccccccc@{}}
\toprule
Method     & \multicolumn{2}{c}{Consistency} & \multicolumn{2}{c}{Relevance} & \multicolumn{2}{c}{Coherence}  & \multicolumn{2}{c}{AVG} \\ 
           & $\rho$        & $\rho_s$       & $\rho$       & $\rho_s$          & $\rho$      & $\rho_s$     & $\rho$    & $\rho_s$   \\ \midrule
Reference-guided & 0,53           & 0,54           & 0,54          & 0,55          & 0,54          & 0,58          & 0,56         & 0,58         \\
Candidate-guided & 0,52           & 0,51           & 0,56          & 0,57          & 0,57          & 0,58          & 0,58         & 0,57         \\
F1               & 0,50           & 0,49           & 0,53          & 0,55          & 0,55          & 0,61          & 0,55         & 0,57         \\ \bottomrule
\end{tabular}
\end{table}

%% file: tables/check_eval.tex
\begin{table*}[t]
\centering
\caption{\textsc{Check-Eval} results on \textsc{SummEval}. Summary-level ($\rho_s$) Spearman and ($\tau$) Kendall-Tau correlations.}
\label{tab:check_eval}

\begin{tabular}{@{}lcccccccccc@{}}
\toprule
Method     & \multicolumn{2}{c}{Consistency} & \multicolumn{2}{c}{Relevance} & \multicolumn{2}{c}{Coherence} & \multicolumn{2}{c}{Fluency} & \multicolumn{2}{c}{AVG} \\ 
           & $\rho_s$        & $\tau$       & $\rho_s$       & $\tau$      & $\rho_s$       & $\tau$      & $\rho_s$      & $\tau$     & $\rho_s$    & $\tau$   \\ \midrule
G-eval     & 0.51           & 0.42         & \textbf{0.55}          & \textbf{0.43}        & \textbf{0.58}          & \textbf{0.46}        & 0.45         & 0.38       & 0.51       & 0.42     \\
GPTScore   & 0.45           &               & 0.38          &              & 0.43          &              & 0.40         &             & 0.41       &           \\
\textsc{Check-Eval} & \textbf{0.60}           & \textbf{0.57}         & 0.50         & 0.42       & 0.56          & 0.46        & \textbf{0.49}         & \textbf{0.44}       & \textbf{0.62}       & \textbf{0.49}     \\ \bottomrule
\end{tabular}

\end{table*}